\documentclass{article}
\usepackage{arxiv}
\usepackage[utf8]{inputenc} 
\usepackage[T1]{fontenc}    
\usepackage{hyperref}       
\usepackage{url}            
\usepackage{booktabs}            
\usepackage{nicefrac}       
\usepackage{microtype}      
\usepackage{lipsum}		
\usepackage{graphicx}
\usepackage{natbib}
\usepackage{doi}
\usepackage{multirow}
\usepackage[table,xcdraw]{xcolor}
\usepackage{subcaption}
\usepackage{amsmath,amssymb,amsfonts}
\usepackage{bm}
\usepackage{algorithm}
\usepackage{algpseudocode}
\usepackage{tikz}
\usepackage{verbatim}

\hypersetup{
    colorlinks=true,     
    linkcolor=black,     
    urlcolor=black,      
    citecolor=blue,
}

\title{Multidimensional classification of posts for online course discussion forum curation}
\date{} 
\author{ \href{https://orcid.org/0000-0001-5839-0273}{\includegraphics[scale=0.06]{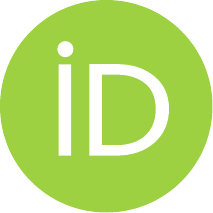}\hspace{1mm}Antonio Leandro Martins Candido} \\
	Federal Institute of Education, Science, and Technology of Ceará (IFCE) - Caucaia, CE, Brazil~ \\
 State University of Ceará -- CCT-PPGCC-UECE -- 60740-903 -- Fortaleza, CE -- Brazil~\\
	\texttt{leandro.candido@aluno.uece.br} \\
	\And
	\href{https://orcid.org/0000-0002-4983-1724}{\includegraphics[scale=0.06]{orcid.pdf}\hspace{1mm}Jos\'e Everardo Bessa ~Maia} \\
        State University of Ceará -- CCT-PPGCC-UECE -- 60740-903 -- Fortaleza, CE -- Brazil~\\
	\texttt{jose.maia@uece.br} \\
}

\renewcommand{\undertitle}{}
\hypersetup{
pdftitle={Multidimensional classification of posts for online course discussion forum curation},
pdfsubject={cs.AI},
pdfauthor={Antonio L. M.~Candido, Jose E. B.~Maia},
pdfkeywords={online course forum curation, multidimensional classification, Bayesian fusion, Large Language Model},
}

\begin{document}
\maketitle

\begin{abstract}
The automatic curation of discussion forums in online courses requires constant updates, making frequent retraining of Large Language Models (LLMs) a resource-intensive process. To circumvent the need for costly fine-tuning, this paper proposes and evaluates the use of Bayesian fusion. The approach combines the multidimensional classification scores of a pre-trained generic LLM with those of a classifier trained on local data. The performance comparison demonstrated that the proposed fusion improves the results compared to each classifier individually, and is competitive with the LLM fine-tuning approach.
\end{abstract}


\section{Introduction}

Multidimensional (or multiple outputs) classification \citep{jia2024multi-dimensional} is characterized by considering the correlation between classes (labels) in addition to the correlation between features and class present in one-dimensional classification. Building an independent classifier for each class is a viable option \citep{candido2023adaptation}, but it does not make use of the correlation between classes, which can often make an important difference. In fact, this information can be considered to improve the decision frontier in ambiguous regions for a one-dimensional classifier.
This is particularly significant in the classification of natural language texts, where language ambiguity is reflected in feature ambiguity and plays a central role in discerning text categories.

\subsection{Context: Curation of online pedagogical forums}

Online pedagogical forum curation refers to the process of selecting, organizing, and contextualizing content and interactions in virtual learning environments, with the aim of promoting meaningful discussions, participant engagement, and collaborative knowledge construction. This process is essential to transforming digital spaces into effective learning communities, especially in distance education and massive online courses (MOOCs) contexts.

In online educational environments, discussion forums play a crucial role in facilitating interaction between students and instructors. However, without proper curation, these forums can become disorganized and less effective in promoting learning. Effective curation involves active moderation, fostering relevant discussions, and integrating relevant educational resources. For example, the MUIR framework proposes cross-linking resources in MOOCs to enrich discussion forums, improving the learning experience of participants \citep{an2020muir}.

In addition, curation in online pedagogical forums also involves creating an environment that supports a community of inquiry, where participants engage in critical and reflective discussions. Guidelines for supporting a community of inquiry through online discussion forums have highlighted the importance of strategies that promote critical thinking and collaboration among students \citep{mudau2023guidelines}.

Figure 1 presents our proposed framework for online discussion forum curation. A curation engine, inserted as a forum member, interacts with the forum by reading posts and responding to the forum's status. The engine responds with resource or action recommendations, or even direct responses via dialogue, based on the combination of two sources of information: local knowledge bases, specific to the course and its context, and support from pre-trained Large Language Models (LLM). Algorithm 1 shows the macro steps of the framework's operation. From this algorithm, it can be seen that if the engine's confidence does not exceed a threshold $th$, it forwards the situational state to a human tutor for resolution. The design goal is to refer less than 2\% of intervention situations to humans.

\begin{algorithm}
\caption{Curation engine macro steps}\label{alg:cap}
\begin{algorithmic}[1]
\Require posts, context, $LKB_i$.
\Ensure response to forum \textbf{OR} refer to human.
\State Read new posts and retrieve context and current state.
\State Create and submit analysis prompt to LLM \textbf{OR} Multi-label classification of new posts.
\State Update state.
\State State update confidence $ > Th$? If not, refer to human.
\State Set post-label-status priority.
\State Create and submit a response generation prompt to LLM \textbf{OR} Retrieve a response from a local knowledge base.
\State Retrieve a response complement from a specific local knowledge base.
\State Compose an intervention response and submit it to the forum.
\end{algorithmic}
\end{algorithm}

Large Language Models (LLMs) pretrained on large language data are known to perform well in open domains but suffer from poor performance in specific domains where specialized language is used \citep{gururangan2020dontstop}. This latter is the case in the context of courses in various fields, requiring fine-tuning of the LLM model to achieve better performance. This makes frequent retraining of Large Language Models (LLMs) a resource-intensive process. To circumvent the need for costly fine-tuning, this paper proposes and evaluates the use of Bayesian fusion. The approach combines the multidimensional classification scores of a pre-trained generic LLM with those of a classifier trained on local data. The performance comparison demonstrated that the proposed fusion improves the results compared to each classifier individually, whether the classifier is trained only in the domain or the pre-trained global LLM, and is competitive with the LLM fine-tuning approach.

Digital curation is also recognized as an essential practice in media and digital literacy education. Exploring curation as a core competency in digital and media literacy education emphasizes the need to teach students curation skills so that they can navigate and critically evaluate the vast amount of information available online \citep{mihailidis2013curation}.

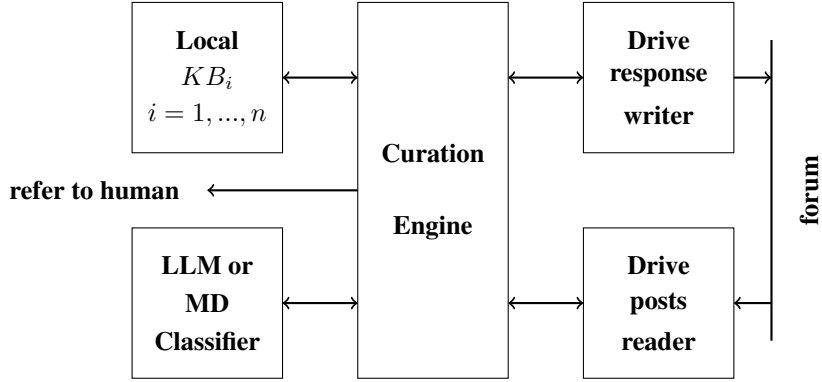
\begin{figure}
   \centering

\begin{tikzpicture}
    \node at (0,0) {};
    \draw (5,0) rectangle (7,5);
    \draw (8,0) rectangle (10,2);
    \draw (2,0) rectangle (4,2);
    \draw (8,3) rectangle (10,5);
    \draw (2,3) rectangle (4,5);
    \draw[thick, ->] (5,2.5) -- (3,2.5);
    \node at (1.5,2.5) {\textbf{refer to human}};
    \draw[thick, ->] (10,4) -- (10.5,4);
    \draw[thick, <-] (10,1) -- (10.5,1);
    \draw[thick, <->] (7,1) -- (8,1);
    \draw[thick, <->] (7,4) -- (8,4);
    \draw[thick, <->] (4,1) -- (5,1);
    \draw[thick, <->] (4,4) -- (5,4);    
    \node at (3,1.5) {\textbf{LLM or}};
    \node at (3,1) {\textbf{MD}};
    \node at (3,0.5) {\textbf{Classifier}};    
    \node at (3,4.5) {\textbf{Local}};
    \node at (3,4) {\textbf{$KB_i$}};
    \node at (3,3.5) {\textbf{$i=1, ..., n$}};    
    \node at (9,1.5) {\textbf{Drive}};
    \node at (9,1) {\textbf{posts}};
    \node at (9,0.5) {\textbf{reader}};    
    \node at (9,4.5) {\textbf{Drive}};
    \node at (9,4) {\textbf{response}};
    \node at (9,3.5) {\textbf{writer}};
    \node at (6,3) {\textbf{Curation}};    
    \node at (6,2) {\textbf{Engine}};
    \node [rotate=90] at (11,2.5) {\textbf{forum}};
    \draw[thick, ] (10.5,0.5) -- (10.5,4.5);
\end{tikzpicture}    
    \caption{Functional structure of the online course forum curation agent software framework.}
    \label{fig:enter-label}
\end{figure}

In summary, curation in online pedagogical forums is a multifaceted practice that involves actively managing content and interactions to promote meaningful and collaborative learning. It requires a combination of pedagogical, technological, and moderation strategies to create effective and engaging online learning environments.

\section{Methods}

\subsection{Multidimensional classification}

Multidimensional classification (MDC) is a supervised learning paradigm in which each input instance is associated with multiple class variables, each representing a distinct semantic dimension. Unlike traditional classification, which associates a single label with each instance, MDC seeks to predict a vector of classes, capturing the semantic complexity of the data. For example, when classifying posts in educational forums, a single post may simultaneously express an opinion, contain a question, and demonstrate urgency, each corresponding to a different dimension \citep{jia2024multi-dimensional}.

MDC differs from multi-label classification mainly in the structure of the outputs. While multi-label classification associates to each instance a set of binary labels coming from a single class space, MDC involves multiple heterogeneous class spaces, each with its own categories. This distinction is crucial because in MDC, the dimensions are semantically distinct and not directly comparable, requiring specific approaches to capture the dependencies between them \citep{jia2024multi-dimensional, huang2015multi-label}.

Modeling the dependencies between the different dimensions is one of the main challenges in MDC. Algorithms such as classifier chains \citep{jia2021decomposition} address this challenge by explicitly modeling the dependencies between class variables. Other approaches, such as the use of deep learning techniques \citep{arya2023deepdependencynetworks}, have been explored to capture these complex relationships more effectively.

MDC has practical applications in a variety of areas, such as online forum curation \citep{zhang2016longitudinal}, where posts need to be classified across multiple dimensions to prioritize responses and interventions. It is also applicable in recommendation systems \citep{oreilly2021recommendation}, sentiment analysis \citep{garg2021drug}, and medical diagnosis \citep{chen2018disease}, where multiple features or symptoms need to be considered simultaneously.

In terms of evaluation, MDC uses metrics that consider accuracy on each dimension individually, as well as aggregate metrics that evaluate the overall performance of the model across all dimensions. These include measures such as micro and macro accuracy, which provide insights into the performance of the model at different levels of granularity \citep{grandini2020metrics, opitz2024evaluation}.

\subsection{Classifier Fusion}

Classifier fusion can be categorized into three levels, based on the type of output each classifier generates. Each level inspires different types of combination rules \citep{xu1992methods}.

\begin{itemize}
\item \textbf{Abstract Level}: Each classifier provides a single class label, and the challenge is to combine these labels into a single final decision. Strategies such as the majority vote rule are common here.

\item \textbf{Rank Level}: The resulting classifiers are ranked from a ranked list of candidate classes, and the goal is to merge these lists to generate a new single classification. The Borda Count method is an example of an approach for this level.

\item \textbf{Measurement Level}: Each classifier assigns a confidence score to each class. The goal is to combine these recommendations to obtain an overall confidence measure. Aggregation operators such as sum, product, and maximum are often used. The approach in this paper falls into the latter category. \end{itemize}

\subsection*{Bayesian fusion}

A common theoretical framework for measurement-level combination rules is the Bayesian approach. In it, suppose that each classifier $C$ estimates the posterior probability $P(\omega_{j}|x)$ that a pattern belongs to a class $\omega_{j}$.

By combining $L$ classifiers, and assuming that they are conditionally statistically independent, the following expression can be derived \citep{kittler1998combining}:

\begin{equation}
P(\omega_{j}|x_{1}, \dots, x_{L}) = \frac{P(\omega_{j})\prod_{l}P(x_{l}|\omega_{j})}{P(x_{1}, \dots, x_{L})} \propto \prod_{l}C_{j,l}(x_{l})
\end{equation}

This formula leads directly to the \textbf{product rule}. With additional assumptions about the distribution of classifiers, other rules such as sum, maximum, minimum, and median can also be justified within this framework.

Unlike traditional ensemble methods such as bagging or boosting, which typically treat models as black boxes and do not explicitly quantify uncertainty, Bayesian fusion allows combining probabilistic predictions from different models, weighting them based on their reliability and variance. This results in a posterior estimate that reflects both the consensus and the degree of confidence of each source. For example, by combining classifiers with different levels of correlation, Bayesian hierarchical models can capture explicit dependencies between them, as demonstrated by \citep{trick2022bayesian}.

One of the main advantages of Bayesian fusion is its ability to handle heterogeneous sources of information, even when they present different levels of noise, granularity or formats. Sander and Beyerer (2012) highlight that, by transforming these sources into a common probabilistic representation, it is possible to perform fusion in a robust manner, even in scenarios of high complexity or uncertainty. Furthermore, the Bayesian approach provides a unified framework for integrating observational data and expert knowledge, facilitating informed decision-making \citep{sander2012bayesian}.

In practical applications, such as recommender systems, sensor fusion or curation of online educational forums, Bayesian fusion can be used to combine different classifiers or heuristics, resulting in a more accurate and adaptive classification. For example, in contexts where multiple models assess the relevance or urgency of forum posts, Bayesian fusion can integrate these assessments, considering the reliability of each model and providing a consolidated estimate with associated uncertainty measures.

The applicability of the Bayesian framework is restricted by theoretical and practical limitations. Notably, the assumption of conditional independence between classifiers constitutes a significant simplification, whose validity is questionable. Evidence shows that, contrary to this assumption, correlation between classifiers can, in specific scenarios, be advantageous for overall performance \citep{terrades2009optimal}.

\section{Related works}

Several studies have analyzed the performance of classic classification systems, such as Naive Bayes and SVM (Support Vector Machine), in different scenarios and preprocessing methods. The results indicate that these models have severe limitations \citep{lima2018topical, candido2019moderacao, candido2025rasa, candido2025social}.

Multidimensional classification (MDC) has gained prominence in the machine learning community due to its ability to handle instances associated with multiple class variables, each representing a distinct semantic dimension. Jia and Zhang (2024) provide a comprehensive review of the MDC paradigm, discussing representative algorithms and the challenges associated with modeling dependencies between classes \citep{jia2024multi-dimensional}.

In the context of online pedagogical forums, effective curation is essential to foster meaningful discussions and participant engagement. An et al. (2020) propose the MUIR framework, which aims to enrich discussion forums in MOOCs through cross-linking of resources, improving the learning experience of users \citep{an2020muir}.

In addition, digital curation is recognized as a core competency in media education and digital literacy. Mihailidis and Cohen (2013) explore the importance of teaching curation skills to students, enabling them to navigate and critically evaluate the vast amount of information available online \citep{mihailidis2013curation}. These works highlight the intersection between advanced classification techniques and effective pedagogical practices, underscoring the importance of multidimensional approaches in the analysis and curation of online educational forums.

Bayesian fusion has been widely explored to improve text classification accuracy, especially in contexts with noisy or sparse data. Zhang (2006) proposed a framework based on Genetic Programming to combine evidence from multiple sources, demonstrating significant improvements in the classification of documents with limited or noisy textual content \citep{zhang2006intelligent}.

Wang et al. (2021) introduced the class-specified topic model (CSTM), which assumes the existence of latent topics shared between classes and specific topics for each class. Using Bayesian inference, the model showed superior performance in text classification and summarization tasks, especially in semi-supervised scenarios \citep{wang2021bayesian}.

Linghu et al. (2022) proposed a Bayesian evidential learning approach for few-shot classification, notably by explicitly modeling uncertainty and improving generalization in scenarios with limited data \citep{linghu2022bayesian}.

With the advancement of large-scale language models (LLMs), several approaches have been proposed to apply these technologies to the classification of educational texts. Liu et al. (2024) developed the AGKA (Annotation Guidelines-Based Knowledge Augmentation) method, which uses GPT-4 to extract knowledge from annotation guidelines, improving the classification of engagement in educational texts without the need for extensive fine-tuning \citep{liu2024annotation}.

Koufakou (2023) conducted a comparative study using models such as BERT, RoBERTa, and XLNet for sentiment analysis and thematic classification in course evaluations. The results showed that RoBERTa achieved 95.5\% accuracy in sentiment analysis, while SVM was more effective in topic classification \citep{koufakou2023deep}.

Toba et al. (2024) introduced the BE-Sent method, which performs hierarchical classification by combining sentiment analysis and Bloom's taxonomy in course discussion forums, allowing a deeper evaluation of educational interactions \citep{toba2024bloom}.

Wang et al. (2024) proposed an adaptable and reliable paradigm for text classification using LLMs, demonstrating that few-shot and fine-tuning strategies can outperform traditional methods in several classification tasks \citep{wang2024adaptable}.

\section{Results and analysis}
The developed techniques are tested first for the English language in this paper, and will be developed for the Portuguese (Brazil) language in a future work.

\subsection{Dataset}
We utilized the Stanford MOOC forum post dataset\footnote[1]{http://datastage.stanford.edu/StanfordMoocPosts}, which comprises 29,604 anonymized posts gathered from 11 distinct online courses. This dataset is accessible to academic researchers upon request. The courses span three primary fields: education, humanities/sciences, and medicine. Each entry in the dataset was hand-labeled by three independent annotators to establish a gold-standard reference \citep{agrawal2015youedu}.

Each post was evaluated across six dimensions: opinion, question, answer, sentiment, confusion, and urgency. The opinion, question, and answer categories were assigned binary labels, while sentiment, confusion, and urgency were rated on a 1–7 scale to facilitate exploration of the following research questions:

\begin{itemize}
    \item Is this post a question? (Yes/No)
    \item Is this post an answer? (Yes/No)
    \item Is this post an opinion? (Yes/No)
    \item What sentiment does this post express? (1-7)
    \item How much confusion does this post express? (1-7)
    \item How much urgent intervention in this post? (1-7)
\end{itemize}

To clarify the urgency scale: a post rated as 1 indicates no urgency and can be safely ignored, while a rating of 7 signifies a highly urgent message that requires immediate attention from an instructor. For the purpose of binary classification, we transformed the 1–7 urgency scale into two classes by labeling posts with urgency scores below 4 as 0 (non-urgent), and those with scores of 4 or higher as 1 (urgent). Additionally, we excluded posts that contained only numeric characters, resulting in a final dataset of 29,604 posts. Among these, 23,186 were categorized as non-urgent, and 6,418 as urgent. Further details on post distribution can be found in Table 1.

\begin{table}[ht]
\centering
\begin{tabular}{l|l|l|l|l|l|l|l}
\hline
areas / sets & no / yes & opinion & question & answer & confusion & sentiment & urgency \\ \hline
DS1 & no & 938 & 9168 & 9690 & 6715 & 1690 & 9418 \\
 & yes & 8941 & 711 & 189 & 3164 & 8189 & 461 \\ \hline
DS2 & no & 2782 & 3574 & 3179 & 481 & 1277 & 3314 \\
 & yes & 2402 & 1610 & 2005 & 4703 & 3907 & 1870 \\ \hline
DS3 & no & 2471 & 2232 & 2244 & 490 & 196 & 2053 \\
 & yes & 559 & 798 & 786 & 2540 & 2834 & 977 \\ \hline
Education & no & 938 & 9168 & 9690 & 6715 & 1690 & 9418 \\
 & yes & 8941 & 711 & 189 & 3164 & 8189 & 461 \\ \hline
\multirow{2}{*}{\begin{tabular}[c]{@{}l@{}}Humanities / \\ Science\end{tabular}} & no & 6016 & 7671 & 8002 & 1358 & 846 & 7247 \\
 & yes & 3707 & 2052 & 1721 & 8365 & 8877 & 2476 \\ \hline
Medicine & no & 6181 & 6806 & 5857 & 1581 & 1851 & 6521 \\
 & yes & 3821 & 3196 & 4145 & 8421 & 8151 & 3481 \\ \hline
\end{tabular}
\caption{Distribution of samples in their respective domains. The posts from the StanfordMOOC dataset are divided into groups. The first groups, DS1 (Education – EDUC115N: How to Learn Math), DS2 (Medicine – SciWrite, Fall 2013), and DS3 (Humanities/Science – Statistical Learning, Winter 2014). The second groups, posts by subject area Education (EDU), Medicine (MED), and Humanities \& Sciences (H\&S).}
\label{tab:dataset}
\end{table}

\subsection{Results}

\begin{table}[ht]
    \small
    \centering
    \begin{tabular}{c|c|c|c|c|c|c|c|c|c|c|c|c}
    & \multicolumn{3}{c|}{MD classif.}&\multicolumn{3}{c|}{LLM GPT}&\multicolumn{3}{c|}{Bayesian Fusion}&\multicolumn{3}{c}{LLM fine-tuning}\\
    \hline
       Configuration  & P & R & F1 & P & R & F1 & P & R & F1 & P & R & F1 \\
         \hline
        Intracourse & 0.81 & 0.80 & 0.78 & 0.80 & 0.78 & 0.77 & 0.81 & 0.80 & 0.78 & 0.82 & 0.82 & 0.80\\
         \hline
        Intradomain & 0.79 & 0.79 & 0.77 & 0.84 & 0.78 & 0.78 & 0.84 & 0.79 & 0.78 & 0.85 & 0.81 & 0.80\\
        \hline
        Crossdomain & 0.73 & 0.68 & 0.67 & - & - & - & 0.73 & 0.68 & 0.67 & 0.77 & 0.70 & 0.69
    \end{tabular}
    \caption{Performance of the multidimensional classification. The average of the test sets for all classes.}
    \label{tab:result1bracis}
\end{table}

Table 2 presents the performance results of multidimensional classification (MD Classif.), generic LLM (GPT), Bayesian fusion, and the LLM fine-tuning approach in three distinct experimental settings: within-course, within-domain, and across-domain. These scenarios were designed to simulate varying degrees of familiarity between the training and testing data, allowing for a robust assessment of the generalizability and robustness of the models. The models were evaluated on a subset of the Stanford MOOC dataset, which includes forum posts from courses in the fields of education, humanities/sciences, and medicine.

In the within-course scenario, where training and testing occur within the same course, the models performed best. Bayesian fusion achieved an F1-score of 0.78, tying with the MD classifier (also with an F1-score of 0.78). The fine-tuned LLM achieved the highest F1-score, with 0.80, while generic GPT achieved 0.77. These results suggest that when the source and target data are identical, specialized approaches such as the MD classifier that exploits correlations between course-specific labels are highly effective.

In the intra-domain setting, where the test data belongs to the same domain (e.g., medicine) but to different courses than the training ones, an interesting change in performance is observed. The fine-tuned LLM again achieved the highest F1-score (0.80), with the highest recall (R) (0.81). However, Bayesian fusion and generic GPT excelled in precision (P), both with 0.84, indicating a lower false positive rate. The F1-score for both was 0.78, outperforming the MD classifier (0.77). This indicates that the generalization ability of the pre-trained LLM starts to become an important differentiator when the specific course context is changed.

The crossdomain scenario is the most challenging, using test data from a completely different domain than the training one. As expected, all models suffer significant performance losses in this scenario. The MD classifier, for example, drops to an F1-score of 0.67, and the fine-tuned LLM to 0.69. Generic GPT was not tested in this setting. Bayesian fusion, with an F1-score of 0.67 and the highest accuracy among the models (0.77), proved to be a more robust approach to deal with the lack of familiarity between domains.

The comparative analysis reveals that there is no single absolute winner in all scenarios. However, the Bayesian fusion approach consistently proves to be strong, matching or outperforming the other approaches, especially in the most challenging scenarios. The results reinforce the central proposition of the paper: fusion is a viable alternative that combines the strengths of different models to improve overall performance, without incurring the high computational cost of full fine-tuning. The final analysis suggests that fine-tuning of LLM, despite its high scores, may not justify the cost relative to the gain obtained in contexts that require rapid adaptation.

\subsection{Analysis}

The comparative analysis of the results reveals a clear dynamic between the model’s specialization and its generalization capacity. The MD classifier, trained with local data, stands out in the intra-course scenario with an F1-score of 0.78. In this context, its ability to efficiently explore correlations between course-specific labels justifies its high performance. However, its performance progressively declines as the context moves away from the original training, falling to an F1-score of 0.77 in the intra-domain scenario and plummeting to 0.67 in the crossdomain scenario. This highlights its strong dependence on specific contextual data, limiting its robustness in broader environments.

On the other hand, the generic LLM (GPT) demonstrates a remarkable generalization capacity, especially in the intra-domain scenario, where it achieves an F1-score of 0.78 with an accuracy of 0.84. Its high accuracy suggests a lower false positive rate, which is extremely valuable in a curation system to avoid incorrect interventions. However, the LLM fine-tuning approach, despite achieving an F1-score of 0.80 in both intra-course and intra-domain scenarios, does not justify the high computational cost. The study itself raises questions about its effectiveness in fast-adaptation contexts, since the performance gain may not compensate for the training effort. 

In this context, Bayesian fusion emerges as the most balanced and robust solution. By combining the confidence scores of a generic LLM with those of a locally trained classifier, the fusion achieves competitive and stable performance in all scenarios, with F1-scores of 0.78 (intra-course), 0.78 (intra-domain), and 0.67 (crossdomain). In the most challenging scenario (crossdomain), Bayesian fusion matches the performance of the MD classifier, but with higher and more balanced precision and recall. This result is significant, as it shows that the fusion can inherit the generalization capacity of the LLM and, at the same time, benefit from the specialization of the local classifier, creating a consolidated and reliable estimate. The approach aligns perfectly with the project’s motivation: to avoid frequent and costly retraining of LLMs by offering an efficient and high-performance alternative that combines the strengths of different models.

\noindent\textbf{Overall}: Referring to Table 7, fine-tuning with the domain-specific knowledge base was expected to produce a greater difference in LLM performance. Looking at the F1 score columns, we can see that, for this dataset, the impact was small. We intend to investigate why this happens and how to build local knowledge bases with a greater impact on LLM performance.

\section{Conclusion}

This work proposed and evaluated the use of Bayesian fusion to combine the multidimensional classification of a generic LLM with a classifier trained on local data, aiming at the curation of discussion forums in online courses. The main motivation was to find an alternative to the fine-tuning process of LLMs, which, although powerful, is resource-intensive and impractical for contexts such as educational forums, where frequent retraining would be necessary. The complexity of the task is accentuated by the nature of multidimensional classification, in which a single post can simultaneously express an opinion, contain a question and demonstrate urgency, requiring sophisticated analysis.

The results demonstrated that the Bayesian fusion approach is not only feasible, but highly competitive. The fusion managed to improve the performance compared to each classifier acting individually and proved to be a robust alternative, with performance competitive to that of fine-tuning LLMs. In scenarios of high specificity (intracourse) and generalization (intradomain and crossdomain), Bayesian fusion maintained a consistently high and balanced performance, standing out as the most robust approach among those analyzed. The ability to maintain solid performance even in the crossdomain scenario, the most challenging, highlights its effectiveness in environments with heterogeneous data.

It is concluded that Bayesian fusion represents an effective and efficient strategy for the curation of educational forums. The technique capitalizes on the best of both worlds: it takes advantage of the extensive knowledge of pre-trained LLMs and the expertise of local models, providing an accurate and adaptable solution. In this way, the study demonstrates a practical and sustainable path for the implementation of artificial intelligence tools in educational environments, bypassing the barrier of high computational cost associated with the complete retraining of large language models, without sacrificing classification quality.

\bibliographystyle{unsrtnat}
\bibliography{references}
\end{document}